\documentclass[letterpaper]{article} 
\usepackage{aaai24}  
\usepackage{times}  
\usepackage{helvet}  
\usepackage{courier}  
\usepackage[hyphens]{url}  
\usepackage{graphicx} 
\usepackage{natbib}  
\usepackage{caption} 
\usepackage{algorithm}
\usepackage{algorithmic}

\usepackage{amssymb}
\usepackage{amsmath}
\usepackage{booktabs}
\usepackage{multirow}
\usepackage{cite}
\usepackage{newfloat}
\usepackage{listings}
\DeclareCaptionStyle{ruled}{labelfont=normalfont,labelsep=colon,strut=off} 
\floatstyle{ruled}
\newfloat{listing}{tb}{lst}{}
\floatname{listing}{Listing}
\title{Mimic: Speaking Style Disentanglement for Speech-Driven 3D Facial Animation}
\author{
    Hui Fu\textsuperscript{\rm 1, \rm3},
    Zeqing Wang\textsuperscript{\rm 2},
    Ke Gong\textsuperscript{\rm 3},
    Keze Wang\textsuperscript{\rm 2},
    Tianshui Chen\textsuperscript{\rm 4},
    Haojie Li\textsuperscript{\rm 3},
    Haifeng Zeng\textsuperscript{\rm 3},
    Wenxiong Kang\textsuperscript{\rm 1}\thanks{Corresponding author}
}
\affiliations{
    \textsuperscript{\rm 1}South China University of Technology,
    \textsuperscript{\rm 2}Sun Yat-sen University,
    \textsuperscript{\rm 3}X-ERA.ai,
    \textsuperscript{\rm 4}Guangdong University of Technology \\

    au\_fuhui@mail.scut.edu.cn, wangzq73@mail2.sysu.edu.cn,\\
    \{kegong936, kezewang, tianshuichen, 12hjli4, haifengzeng459\}@gamil.com, auwxkang@scut.edu.cn
}

\usepackage{bibentry}

\begin{document}

\maketitle

\begin{abstract}
Speech-driven 3D facial animation aims to synthesize vivid facial animations that accurately synchronize with speech and match the unique speaking style. However, existing works primarily focus on achieving precise lip synchronization while neglecting to model the subject-specific speaking style, often resulting in unrealistic facial animations. To the best of our knowledge, this work makes the first attempt to explore the coupled information between the speaking style and the semantic content in facial motions. Specifically, we introduce an innovative speaking style disentanglement method, which enables arbitrary-subject speaking style encoding and leads to a more realistic synthesis of speech-driven facial animations. Subsequently, we propose a novel framework called \textbf{Mimic} to learn disentangled representations of the speaking style and content from facial motions by building two latent spaces for style and content, respectively. Moreover, to facilitate disentangled representation learning, we introduce four well-designed constraints: an auxiliary style classifier, an auxiliary inverse classifier, a content contrastive loss, and a pair of latent cycle losses, which can effectively contribute to the construction of the identity-related style space and semantic-related content space. Extensive qualitative and quantitative experiments conducted on three publicly available datasets demonstrate that our approach outperforms state-of-the-art methods and is capable of capturing diverse speaking styles for speech-driven 3D facial animation. The source code and supplementary video are publicly available at: https://zeqing-wang.github.io/Mimic/
\end{abstract}

\section{Introduction}
\begin{figure}[ht]
    \centering
    \includegraphics[width=0.5\textwidth]
    {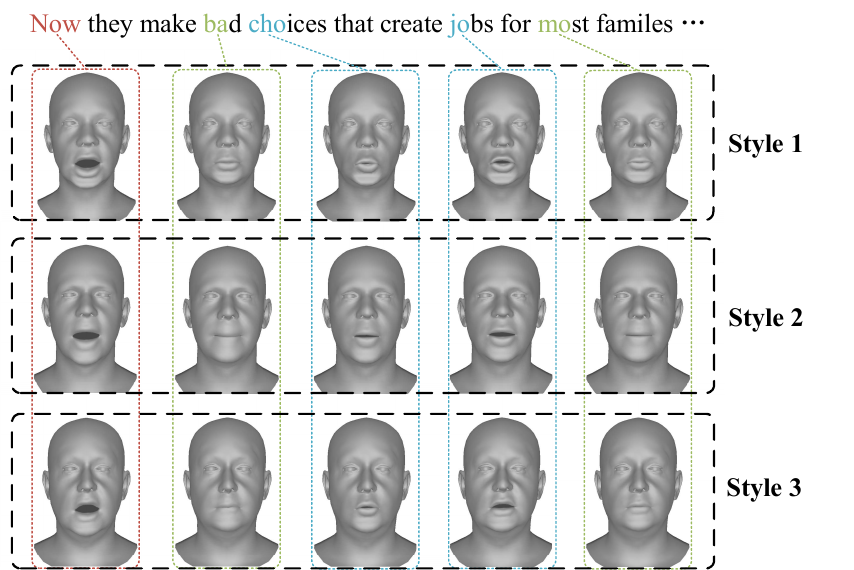}
    \caption{Illustrations of various speaking styles, which reveals the differences in the amplitudes of mouth opening (red border) and closing (green border), as well as the level of dimensionality in pouting (blue border).}
    \label{introduction}
\end{figure}

Speech-driven 3D facial animation is receiving increasing attention due to its broad applications, such as digital avatar synthesis, movie production, etc. Recently, it has become possible to generate fluent and lip-synchronized facial animations with the advent of deep learning. However, most previous works focus only on learning the speaking content, going for synchronization of speech and lip movements \cite{voca,meshtalk,faceformer,codetalker}. Learning speaking style has often been neglected. 
The speaking style reflects the unique lip movement pattern of a speaker, including the amplitude of mouth opening and closing, the dimensionality of pouting, etc. In real-world scenarios, individuals exhibit remarkably distinct speaking styles even when delivering the same speech, as depicted in Figure \ref{introduction}. 
These significant diversities in speaking styles pose a challenge in producing facial animations that accurately match an individual's unique speaking style.

Previous works usually utilize a person-specific model or treat speaking styles simply as identity classes. The former approaches \cite{karras2017audio,richard2021audio} can generate facial animations for a specific individual with a corresponding speaking style but lack generalization ability across diverse subjects. The latter methods \cite{voca,meshtalk,faceformer,codetalker} employ a one-hot encoding of the training identities to distinguish different speaking styles and can control the synthesized speaking style by modifying the one-hot vector. However, such a formulation is far from representing informative speaking styles. Besides, it fails to match the new speaking style of an unseen subject in the training stage. 
To solve this, Imitator \cite{imitator} aims at designing a style embedding layer with one-hot encoding to optimize for identity-specific speaking style based on a reference video. Nevertheless, it depends on a time-consuming two-stage adaptation, leading to a dramatic performance drop when only a few frames are available for adaptation.

Unlike the mainstream approaches that still rely on one-hot vectors to represent style information, we make the first effort to mine the inherent nature of speaking style manifested in facial animations.
We begin by exploring the motion characteristics of facial animations through two experiments. Specifically, we train a classifier to categorize the facial motion sequences of different identities and a recognizer to convert facial motion sequences into text, which results in high test classification accuracy and low test word error rate, respectively, as presented in Table \ref{motivation}. The results suggest that the facial motion sequence encompasses both identity-related speaking style information and semantic-related content information. 
The coupled speaking style and content in facial motions make it challenging to model stylized facial animations in a purely data-driven manner.

\begin{table}[]
    \centering
    \scalebox{0.7}{
        \begin{tabular}{@{}lccc@{}}
        \toprule
        Task          & Data           & Classification accuracy & Word error rate \\ \midrule
        Style classification & 150 identities & 89.1\%                  & ——              \\
        Content recognition  & 2896 sentences & ——                      & 32.6\%            \\ \bottomrule
        \end{tabular}
    }
    \caption{Results of our exploratory experiments.}
    \label{motivation}
\end{table}

In this paper, to optimize the speaking style in speech-driven 3D facial animation by disentangling it from the facial motions, we propose a novel disentanglement framework called \textbf{Mimic}, which learns disentangled representations of speaking style and content from facial motions by constructing two separated latent spaces. 
It begins by encoding a facial motion sequence into the style space and content space using the elaborately designed style encoder and content encoder. Then, a synchronized speech clip is encoded into the content space by an audio encoder. Last, a motion decoder is employed to reconstruct the motion sequence with the specific speaking style by combining two latent codes from the style space and the content space, respectively. 
During inference, our style encoder can map an arbitrary facial motion sequence into the style space to obtain the corresponding style code. Combining the style code with the audio features extracted from the driving speech, the motion decoder can generate stylized facial motions and thus produce facial animations matching the identity-specific speaking style by adding a template mesh.


However, directly constructing the two disentangled latent spaces is not a trivial task due to the significant variations of facial motions considering the diversity of potential subjects and speeches. To solve this, we further design four effective constraints for better disentangling of style and content. First, to facilitate our style encoder to encode identity-specific style representations, we utilize an auxiliary style classifier to cluster the style codes of the same identity. Second, to reduce the overlap between the content space and style space, we apply an auxiliary inverse classifier with a gradient reverse layer (GRL) \cite{grl} to the content codes to eliminate the identity-related information from the content space. Third, as the content space is expected to contain semantic content information, we adopt a content contrastive loss \cite{clip} to align the content codes and audio features, enabling the injection of speech-related semantic information into the content space. Last, to establish semantically meaningful latent spaces, we introduce a pair of latent cycle losses to enhance the representational ability of the latent codes. By incorporating the four constraints, we successfully build an identity-related style space and a semantic-related content space.
In summary, our main contributions are as follows:
\begin{itemize}
    \item We make the first effort to mine the speaking style in facial animations. By experimentally demonstrating the coupled speaking style and content in facial motions, we propose to disentangle and optimize the speaking style to boost the speech-driven 3D facial animation task.
    \item We introduce a novel framework with four elaborate constraints that can effectively learn disentangled representations of the speaking style and content from facial motions by building two latent spaces for style and content. As a result, we achieve arbitrary-subject speaking style encoding and successfully produce authentic 3D facial animations matching the identity-specific speaking style.
    \item We conduct extensive qualitative and quantitative experiments on three public datasets, demonstrating our method outperforms existing state-of-the-art methods.
\end{itemize}
    

\begin{figure*}[ht]
    \centering
    \includegraphics[width=0.85\textwidth]
    {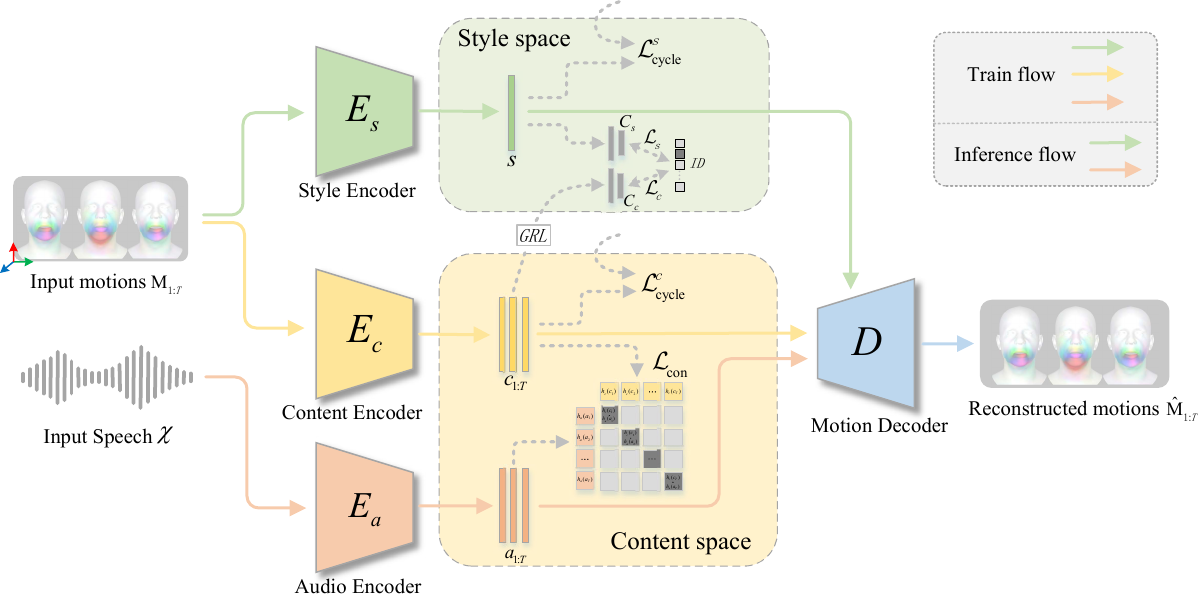}
    \caption{
    Illustration of Mimic that learns two disentangled latent spaces for style and content, respectively. The input facial motion sequence $\mathbf{M}_{1:T}$ are encoded into the two spaces by the style encoder $E_s$ and content encoder $E_c$, obtaining a style code $s$ and sequential content codes $c_{1:T}$. The synchronized speech $\mathcal{X}$ is encoded into audio features $a_{1:T}$ by the audio encoder $E_a$. The motion decoder $D$ generates facial motions $\mathbf{\hat{M}}_{1:T}$ by combining $s$ with either $c_{1:T}$ or $a_{1:T}$. During inference, the style encoder can encode a short style reference motion sequence of the target subject to get a style code. Together with the audio features extracted from the driving speech, we can produce facial animations matching an identity-specific speaking style.
    }
    \label{framework}
\end{figure*}

\section{Related Work}
\subsubsection{Speech-Driven 3D Facial Animation}
Previous works for speech-driven 3D facial animation can be roughly categorized into parameter-based and vertex-based. Parameter-based methods \cite{jali,visemenet,learningviseme} usually establish a set of mapping rules between phonemes and their visual counterparts, e.g., visemes, at the cost of lots of complex procedures. We focus on vertex-based methods, which aim to directly learn the mapping from speech to mesh vertices. VOCA \cite{voca} models multi-subject vertex animation with a one-hot identity encoding. MeshTalk \cite{meshtalk} further extends the work from lower-face to whole-face animation. FaceFormer \cite{faceformer} considers the long-term audio context by a transformer-based model. CodeTalker \cite{codetalker} models the facial motion space with discrete primitives and achieves expressive facial animations. However, all these methods use a one-hot encoding of training identities to distinguish speaking styles, making it fail to match new speaking styles. Recent Imitator \cite{imitator} proposes a method that can adapt to new subjects. It designs a two-stage adaptation strategy to learn the new speaking style of the unseen speaker based on a reference video, which is time-consuming and resource-consuming. Furthermore, its performance drops dramatically when there are only a few frames for adaptation. In this work, We aim at a method that can encode the new speaking style of an unseen subject without any adaptation and perform well with only a few reference frames.

\subsubsection{Disentanglement in Talking Face Generation}
Disentanglement has been widely studied in talking face generation recently. Several works focus on disentangling the emotion-related information from the talking face or speech. \citet{EAMM} design a special data augmentation strategy to detach emotion from other factors. \citet{DPE} introduce a bidirectional cyclic training strategy to decouple pose and expression. \citet{emotalk} propose a framework to disentangle the emotion and content in the speech. A few studies pay attention to identity-related information. \citet{makelttalk} disentangle the content and identity information in the input audio signal, while \citet{zhou2019talking} learn to disentangle the identity-related and speech-related information from 2D talking face by adversarial training. In this work, we aim to disentangle the identity-related speaking style and semantic-related content in 3D facial motions.

\section{Proposed Method}
\subsection{Motivation}\label{Motivation}
\label{sec_motivation}
To exploit the speaking style in facial animation, we conducted two exploratory experiments based on facial motions that represent 3D vertex displacements on top of a template mesh. The data are from our built 3D-HDTF dataset. First, we train a classifier to classify the facial motion sequences of different identities. It consists of several temporal convolutional layers and a multi-layer transformer encoder followed by a temporal pooling layer. We use 150 identities with 210 videos, where 20\% of frames per video are reserved for testing. As reported in Table \ref{motivation}, the test accuracy reaches an impressive value of 89.1\%, suggesting there exists identity-related speaking style information in the facial motions. Second, we develop a recognizer to convert the facial motion sequence into text. The recognizer is trained using the CTC loss \cite{ctcloss} and consists of a motion encoder and a text decoder, where the motion encoder is similar to the classifier but without the last temporal pooling layer, while the text decoder is achieved by a fully connected layer. We use 2316 sentences for training and an additional 580 sentences for testing. We test the word error rate (WER) and report a WER of 32.6\% (Table \ref{motivation}), indicating the presence of semantic content information within the facial motions. The above results suggest that facial motions contain coupled speaking style and semantic content, which makes it challenging to directly model stylized facial animations. 
To this end, we propose to optimize the speaking style by disentangling it from the facial motions.

\subsection{Overview}
We propose Mimic for style-content disentanglement and synthesizing facial animations matching an identity-specific speaking style, as illustrated in Figure \ref{framework}. 
During training, we aim to learn disentangled style space and content space with the input motion sequence $\mathbf{M}_{1: T}$ of $T$ frames and a synchronized speech $\mathcal{X}$. Our Mimic consists of four modules: (1) a style encoder $E_s$ encodes the $\mathbf{M}_{1: T}$ into a compact style code $s$; (2) a content encoder $E_c$ encodes the $\mathbf{M}_{1: T}$ into sequential content codes $c_{1:T}$; (3) an audio encoder $E_a$ extracts the audio features $a_{1:T}$ from $\mathcal{X}$. (4) a motion decoder $D$ generates facial motions $\mathbf{\hat{M}}_{1: T}$ by combining $s$ with either $a_{1:T}$ or $c_{1:T}$. 
To ensure effective learning of disentangled latent spaces, we incorporate four well-designed constraints, which will be described in the following sections.

\subsection{Architecture}
\subsubsection{Style Encoder} 
The style encoder encodes the input facial motion sequence into a compact latent code $s \in \mathbb{R}^{d_{s}}$ that captures identity-specific speaking style.
The speaking style exists in the time series but remains constant irrespective of temporal variations. Here we first employ a temporal convolutions network (TCN) to transform the input motions $\mathbf{M}_{1:T}$ into style tokens, then utilize a multi-layer transformer encoder to convert the tokens into contextualized style representations. After modeling the temporal correlation, we apply temporal mean pooling to aggregate the style information and finally obtain the compact style code $s$.


\subsubsection{Content Encoder}
The content encoder encodes the input motion sequence into sequential latent codes $c_{1:T}\in\mathbb{R}^{T\times d_{c}}$ capturing the semantic content. Similar to the style encoder, we use a TCN and a transformer encoder to extract content tokens and model contextual information. We replace layer normalization (LN) in TCN with instance normalization (IN) without affine transformation, which has been widely used in image style transfer \cite{huang2017arbitrary} and voice conversion \cite{chou2019one}. 


\subsubsection{Audio Encoder}
The audio encoder is to extract the features $a_{1:T} \in\mathbb{R}^{T \times d_{a}}$ from input speech $\mathcal{X}$, which injects the semantic information into the content space.
We follow a self-supervised pre-trained speech model, wav2vec 2.0 \cite{wav2vec2.0}. In detail, our audio encoder consists of an audio feature extractor, transformer encoder, and top linear projection layer. We initialized the audio feature extractor and transformer encoder using pre-trained weights from wav2vec 2.0. The feature extractor uses a TCN structure to convert the waveform into speech tokens  $a^{\prime}_{1:T_a}$ with frequency $f_a$. Since the facial motions might be captured with a frequency $f_{m}$ not equal to $f_a$, an alignment step is necessary. Instead of using a linear interpolation layer \cite{faceformer}, we adopt a 1-D convolution to downsample audio features along the temporal axis, which minimizes the loss of information. The kernel size ($\text{k}$), padding size ($\text{p}$), and sliding stride ($\text{s}$) are chosen to satisfy the equation $(f_a-1)\text{s}=f_m+2\text{p}-\text{k}$. After aligning, the resulting audio features have a length of $T=\frac{f_m}{f_a}T_a$.
 
\subsubsection{Motion Decoder}
The motion decoder produces sequential facial motions $\mathbf{\hat{M}}_{1:T}$ by combining the style code with either content codes or audio features. The shared decoder for the content codes and the audio features will force them to share their distributions and pull them toward the common distribution space. 
To capture dependencies within the motion sequence, we utilize an autoregressive transformer decoder. It incorporates causal multi-head self-attention, allowing each frame to attend to previous frames in the context of past motions. Additionally, multi-head cross-attention is employed to align the audio-motion modalities.
To incorporate style information, existing methods simply add \cite{faceformer,codetalker} or concatenate \cite{voca,imitator} their style vector with the input tokens of the decoder, which is only suitable for one-hot encoding containing limited information. Inspired by \cite{saln}, we use a style-adaptive layer normalization (SALN) to incorporate style code into the decoding process. Specifically, we replace the LN in our decoder with SALN, which receives the style code $s$ and produces the gain and bias of the input feature of the normalization layer. Formally,
\begin{equation}
    \text{SALN}(\boldsymbol{h},s)=g(s)\cdot \frac{\boldsymbol{h}-\mu(\boldsymbol{h})}{\sigma(\boldsymbol{h})}+b(s)
\end{equation}
where $\boldsymbol{h}\in \mathbb{R}^{T\times d_h}$ is the input feature of SALN. $\mu(\cdot)$ and $\sigma(\cdot)$ denote the mean and standard deviation of $\boldsymbol{h}$ along the dimension $d_h$. $g(\cdot)$ and $b(\cdot)$ are affine transformations adopted by a fully connected layer that can adaptively scale and shift the normalized $\boldsymbol{h}$ according to the given $s$.

\subsection{Learning Disentangled Latent Spaces}
\subsubsection{Auxiliary Style Classifier}
To enforce the clustering of style codes belonging to the same identity within the style space, we introduce an auxiliary style classifier denoted as $\mathbf{C}_s$. This classifier is specifically designed to categorize style codes $s$ of different subjects using a cross-entropy loss:
\begin{equation}
    \mathcal{L}_{s}=-\sum_{i=1}^{N_s} p_i \log \left(\operatorname{softmax}\left(\mathbf{C}_s \left(s \right) \right)
_i \right)
\end{equation}
where $\mathbf{C}_s$ is our style classifier realized by a trainable weight matrix, $p$ is the one-hot label of the identity classes, and $N_{s}$ is the number of identities in the training set. By integrating $\mathbf{C}_s$, we encourage the distributions of style codes for the same subject to be closer in proximity within the style space.

\subsubsection{Auxiliary Inverse Classifier} 
The content space is expected to eliminate identity-related style information and reduce overlap with the style space. Here we utilize an auxiliary inverse classifier over the content codes. It is composed of a gradient reversal layer (GRL) \cite{grl} and an identity-related classifier $\mathbf{C}_c$ similar to the aforementioned style classifier. The classification loss $\mathcal{L}_{c}$ is:
\begin{equation}
    \mathcal{L}_{c}=-\sum_{i=1}^{N_s} p_i \log \left(\operatorname{softmax}\left(\mathbf{C}_c \left(\overline{c} \right) \right)_i \right)
\end{equation}
where $\overline{c}\in \mathbb{R}^{d_c}$ is the temporal average of content codes.
The gradient of $\mathcal{L}_{c}$ is reversed via GRL by multiplying a negative value before backward propagating to the content encoder, forcing the content encoder to be optimized toward decreasing identity classification accuracy. Thus the identity information can be largely removed from the content space, and the distributions of the two spaces are pushed away.

\subsubsection{Content Contrastive Loss}
The content space is expected to contain semantic content information. Inspired by CLIP \cite{clip}, we adopt a content contrastive loss $\mathcal{L}_{\text{con}}$ to align the semantic audio features and content codes, which can be written as
\begin{equation}
    \mathcal{L}_{\text{con}}=\frac{1}{T} \sum_{i=1}^T\left(\lambda \mathcal{L}_i^{(c \rightarrow a)}+(1-\lambda) \mathcal{L}_i^{(c \rightarrow a)}\right)
\end{equation}
where $\lambda=0.5$. $\mathcal{L}_i^{(a \rightarrow c)}$ and $\mathcal{L}_i^{(c \rightarrow a)}$ represent audio-to-motion and motion-to-audio contrastive loss:
\begin{equation}
    \mathcal{L}_i^{(a \rightarrow c)}=-\log \frac{\exp \left[\left\langle h_a\left(a_i\right), h_c\left(c_i\right)\right\rangle / \tau \right]}{\sum_{k=1}^T \exp \left[\left\langle h_a\left(a_i\right), h_c\left(c_k\right)\right\rangle / \tau \right]} 
\end{equation}
\begin{equation}
    \mathcal{L}_i^{(c \rightarrow a)}=-\log \frac{\exp \left[\left\langle h_c\left(c_i\right), h_a\left(a_i\right)\right\rangle / \tau \right]}{\sum_{k=1}^T \exp \left[\left\langle h_c\left(c_i\right), h_a\left(a_k\right)\right\rangle / \tau \right]}
\end{equation}
where $\left\langle \cdot \right\rangle$ denotes the cosine similarity. $\tau \in \mathbb{R}^+$ is a learnable temperature parameter. $h_a(a)$ and $h_c(c)$ are the hidden states of the content codes and audio features.  
Through contrastive learning, the content codes and audio features at the same time step are tightly aligned while the misalignment is pushed away.

\subsubsection{Paired Latent Cycle Losses}
\begin{figure}[ht]
    \centering
    \includegraphics[width=0.4\textwidth]
    {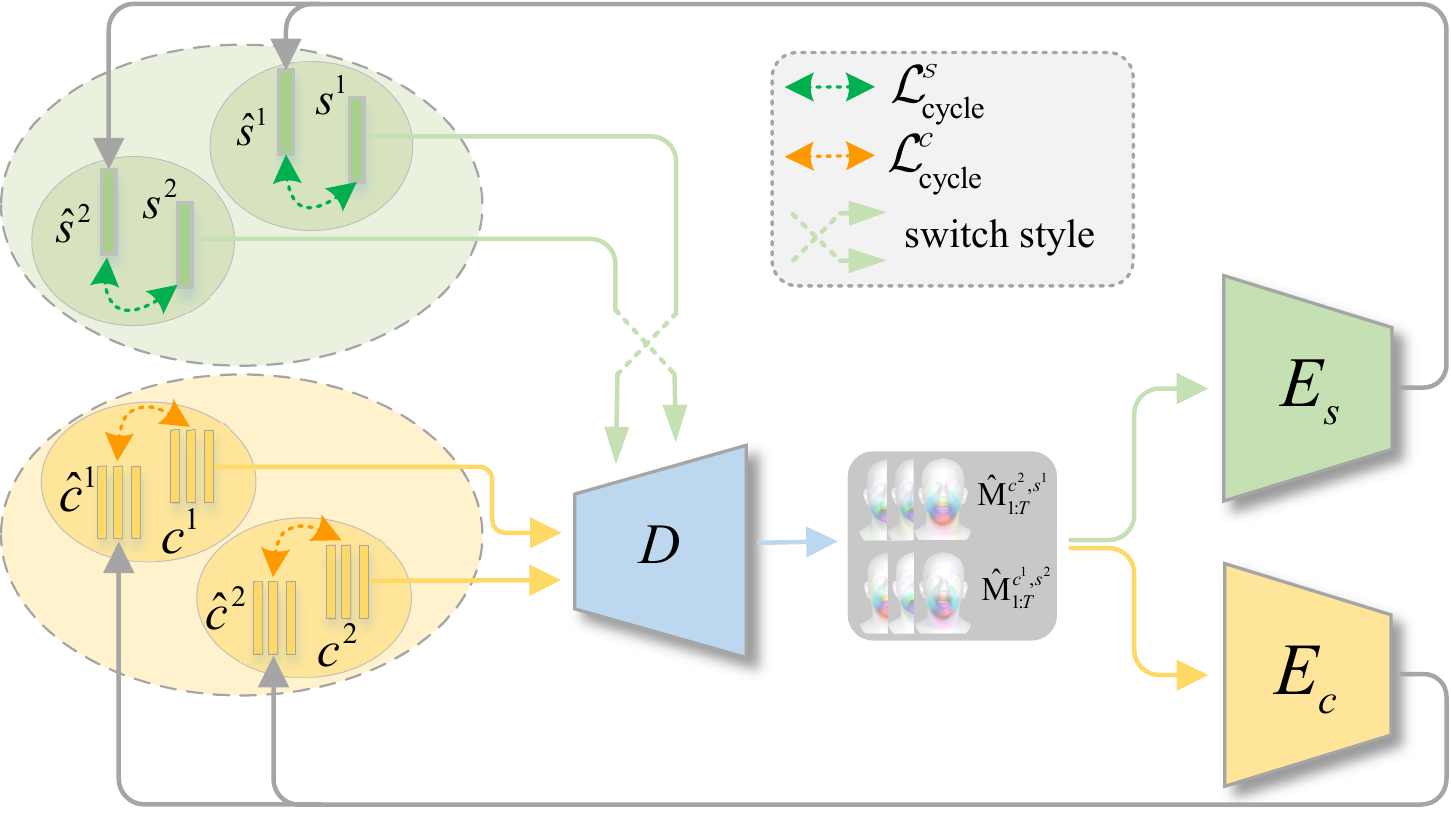}
    \caption{Illustration of latent cycle losses (batch size of 2).}
    \label{latentcycleloss}
\end{figure}

To enhance the representational capabilities of the latent codes, we introduce a pair of latent cycle losses, namely style cycle loss and content cycle loss. As shown in Figure \ref{latentcycleloss}, during training, we first randomly switch the style codes in a mini-batch ($s^1$, $s^2$) (shown by a batch size of 2) and combine the switched style codes ($s^2$, $s^1$) with content codes ($c^1$, $c^2$) to reconstruct the facial motions ($\mathbf{\hat{M}}^{s^2,s^1}_{1:T}$, $\mathbf{\hat{M}}^{s^1,s^2}_{1:T}$), which have no ground-truth reference. Then we respectively employ the style encoder and content encoder to derive cyclic style codes ($\hat{s}^2$, $\hat{s}^1$) and content codes ($\hat{c}^1$, $\hat{c}^2$) from the reconstructed motions. Finally, we compute the cosine similarity loss between cyclic style codes and switched style codes, as well as the contrastive loss between cyclic content codes and raw content codes,
\begin{equation}
    \mathcal{L}_{\text{cycle}}^{s}=\frac{1}{B}\sum_{b=1}^{B}\frac{s^b\cdot \hat{s}^b}{\operatorname{max}(\|s^b\|\cdot \|\hat{s}^b\|, \epsilon)}
\end{equation}
\begin{equation}
    \mathcal{L}_{\text{cycle}}^{c}=\frac{1}{BT}\sum_{b=1}^B \sum_{i=1}^T\left(\lambda \mathcal{L}_i^{(c^b \rightarrow \hat{c}^b)}+(1-\lambda) \mathcal{L}_i^{(\hat{c}^b \rightarrow c^b)}\right)
\end{equation}
Here, $B$ is the batch size, $\mathcal{L}_i^{(c^b \rightarrow \hat{c}^b)}$ and $\mathcal{L}_i^{(\hat{c}^b \rightarrow c^b)}$ are akin to $\mathcal{L}_i^{(a \rightarrow c)}$ and $\mathcal{L}_i^{(c \rightarrow a)}$, and $\epsilon$ is a small value.

\subsubsection{Training Objects}
Our Mimic is trained in a regressive way with a motion regressive loss $\mathcal{L}_{r}$ that measures the vertex-level difference between the reconstructed motions and the ground truth. The loss is defined as:
\begin{equation}
    \mathcal{L}_{r}=\|\mathbf{\hat{M}}_{1: T}^c-\mathbf{M}_{1: T}\|_2+\|\mathbf{\hat{M}}_{1: T}^a-\mathbf{M}_{1: T}\|_2
\end{equation}
where $\mathbf{\hat{M}}_{1: T}^c$ and $\mathbf{\hat{M}}_{1: T}^a$ denote the reconstructed motions generated from the content codes and audio features, respectively.
Our full objective can be written as follows:
\begin{equation}
\begin{aligned}
    \mathcal{L}=\lambda_r\mathcal{L}_{r}+\lambda_s\mathcal{L}_{s}+\lambda_c\mathcal{L}_{c}+\lambda_{\text{con}}\mathcal{L}_{\text{con}} \\
    +\lambda_{\text{cycle}}^s\mathcal{L}_{\text{cycle}}^{s}+\lambda_{\text{cycle}}^c\mathcal{L}_{\text{cycle}
}^{c}
\end{aligned}
\end{equation}
where $\lambda_r=1$, $\lambda_s=2.5\times10^{-7}$, $\lambda_c=5.0\times10^{-7}$, $\lambda_{\text{con}}=5.0\times10^{-7}$, $\lambda_{\text{cycle}}^s=2.5\times10^{-5}$, and $\lambda_{\text{cycle}}^c=5.0\times10^{-6}$.

\section{Experiments}
\subsection{Datasets and Implementations}
\subsubsection{Datasets} \label{datasets}
Previous works \cite{codetalker} usually employ VOCASET \cite{voca} and BIWI \cite{biwi}, which are limited in identities and richness of speech content. In this paper, we conduct a larger dataset called 3D-HDTF based on a high-quality 2D audio-visual dataset HDTF \cite{hdtf}. We generate pseudo ground truth mesh data with the FLAME \cite{flame} topology by integrating an off-the-shelf 3D face reconstruction method named SPECTRE \cite{spectre}. More details can be seen in our supplementary materials. We finally obtained 220 paired audio-mesh sequences of 160 identities with over 3k sentences and corresponding 160 template meshes. We use 172 sequences of 150 identities for training and conduct two testing sets: Test-A contains 38 sequences of 38 seen identities; Test-B contains 10 sequences of 10 unseen identities, all of which have a duration exceeding 4 minutes. We perform quantitative evaluations on 3D-HDTF-Test-A and B, and qualitative evaluations on 3D-HDTF-Test-A and B, VOCA-Test, and BIWI-Test-B. For VOCASET and BIWI, we follow the settings in CodeTalker, illustrated in the supplementary materials.

\subsubsection{Implementation Details}
We use a 6-second window size for both the input sequence during training and the style reference sequence during inference.
We use Adam optimizer with a learning rate of 0.0001 for training. The batch size is set to 6 for 3D-HDTF and 1 for both VOCASET and BIWI. Our framework is implemented by Pytorch, trained on a single RTX 4090 GPU for 150 epochs. 
We compare our method with VOCA, MeshTalk, FaceFormer, CodeTalker, and Imitator. More details of baselines and our Mimic can be seen in the supplementary materials.

\subsection{Quantitative Evaluation}
\subsubsection{Metrics} We adopt five metrics for quantitative evaluation: (1) Face vertex error (FVE), which calculates the L2 vertex distance between the generated face meshes and the ground truth to measure synchronization. (2) Lip vertex error (LVE) \cite{meshtalk}, which is used to measure lip sync. It calculates the maximal L2 error of all lip vertices for each frame and takes the average over all frames in the testing set. (3) Lip Dynamic Time Wrapping (LDTW) \cite{imitator}, which employs Dynamic Time Wrapping in the lip region to evaluate lip sync. (4) Lip dynamics deviation (LDD), which is inspired by upper-face dynamics deviation (FDD) \cite{codetalker} and calculated by replacing the upper-face vertices in FDD with lip vertices. It measures the dynamics property of the lip movements and reflects the speaking style. (5) Style cosine similarity (SCS), which is used to evaluate the speaking style. We first train a facial motion sequence classifier and use it to extract the embeddings of our generated facial motions and the ground truth, then calculate the cosine similarity between them. A higher SCS indicates a more realistic speaking style.

\subsubsection{Comparisons with state-of-the-arts} We calculate the five metrics over all sequences in 3D-HDTF Test-A or Test-B and take the average for comparison. According to Table \ref{quantitativeresults}, our method achieves the best performance among all metrics.
Our method achieves the lowest FVE, LVE, and LDTW, suggesting that it produces the most accurate lip sync. This is due to the fact that we disentangle the speaking style from motions, which encourages the audio encoder to focus on modeling the content.
The best performance in terms of LDD and SCS indicates our method can produce facial animations following the reference speaking style. We achieve better LDD and SCS than other methods on 3D-HDTF-Test-A, demonstrating our style code contains more style information than one-hot encoding.
Our Mimic also outperforms Imitator on 3D-HDTF-Test-B, which shows its stronger generalization ability to unseen subjects.

\begin{table}[]
    \centering
        \scalebox{0.7}{
        \begin{tabular}{@{}lccccc@{}}
        \toprule
        Method     & \begin{tabular}[c]{@{}c@{}}FVE $\downarrow$\\ ( $\times 10^{-6}$\text{mm})\end{tabular} & \begin{tabular}[c]{@{}c@{}}LVE $\downarrow$ \\ ( $\times 10^{-5}$\text{mm})\end{tabular} & \begin{tabular}[c]{@{}c@{}}LDTW $\downarrow$ \\ ($\times 10^{-4}$)\end{tabular} & \begin{tabular}[c]{@{}c@{}}LDD $\downarrow$ \\ ( $\times 10^{-5}$\text{mm})\end{tabular} & SCS $\uparrow$   \\ \midrule
        VOCA       & 1.99                                                                                                                                               & 5.61                                                                                                                                               & 8.28                                                                                                                                      & 2.08                                                                                                                                               & 0.954                           \\
        MeshTalk   & 1.34                                                                                                                                               & 4.91                                                                                                                                               & 8.13                                                                                                                                      & 1.86                                                                                                                                               & 0.956                           \\
        FaceFormer & 1.07                                                                                                                                               & 4.56                                                                                                                                               & 7.79                                                                                                                                      & 1.07                                                                                                                                               & 0.960                           \\
        CodeTalker & 1.02                                                                                                                                               & 4.36                                                                                                                                               & 8.05                                                                                                                                      & 1.03                                                                                                                                               & 0.965                           \\
        \textbf{Ours}       & \textbf{0.55}                                                                                                                     & \textbf{3.20}                                                                                                                     & \textbf{6.82}                                                                                                            & \textbf{0.89}                                                                                                                     & \textbf{0.995} \\ \hline
        Imitator   & 0.85                                                                                                                                               & 6.02                                                                                                                                               & 9.01                                                                                                                                      & 1.01                                                                                                                                               & 0.972                           \\
        \textbf{Ours}       & \textbf{0.60}                                                                                                                     & \textbf{4.05}                                                                                                                     & \textbf{7.43}                                                                                                            & \textbf{0.93}                                                                                                                     & \textbf{0.992} \\ \hline
        \end{tabular}
        }

    \caption{Quantitative evaluations on 3D-HDTF Test-A (top 5 rows) and Test-B (bottom 2 rows).}
    \label{quantitativeresults}
\end{table}

\begin{figure}[ht]
    \centering
    \includegraphics[width=0.45\textwidth]
    {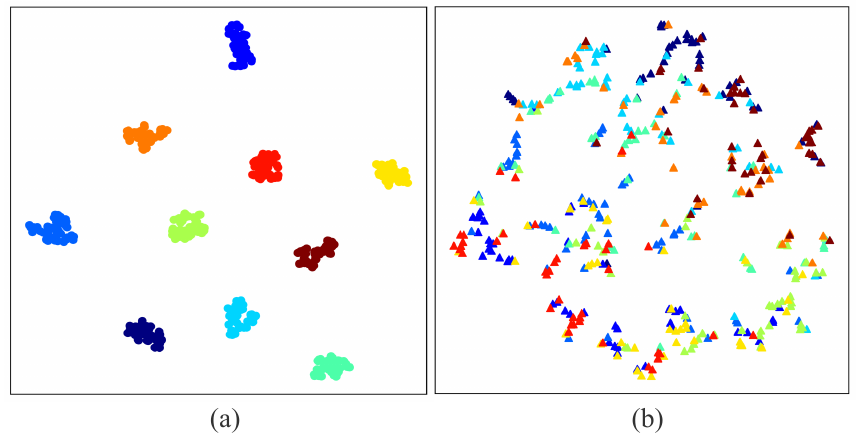}
    \caption{Visualization of style space (a) and content space (b). Different colors for the latent codes of different subjects.}
    \label{latentspace}
\end{figure}

\begin{figure}[ht]
    \centering
    \includegraphics[width=0.45\textwidth]
    {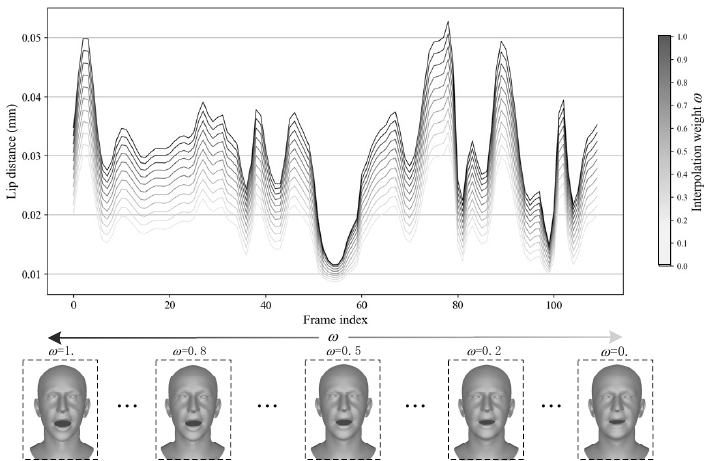}
    \caption{Visualization of style interpolation. The top displays lip distance curves with various interpolation weights, while the bottom shows corresponding facial animations for Frame 75 of the top figure.}
    \label{styleinter}
\end{figure}

\subsection{Qualitative Evaluation}
\subsubsection{Visual Comparisons}
\begin{figure*}[ht]
    \centering
    \includegraphics[width=0.9\textwidth]
    {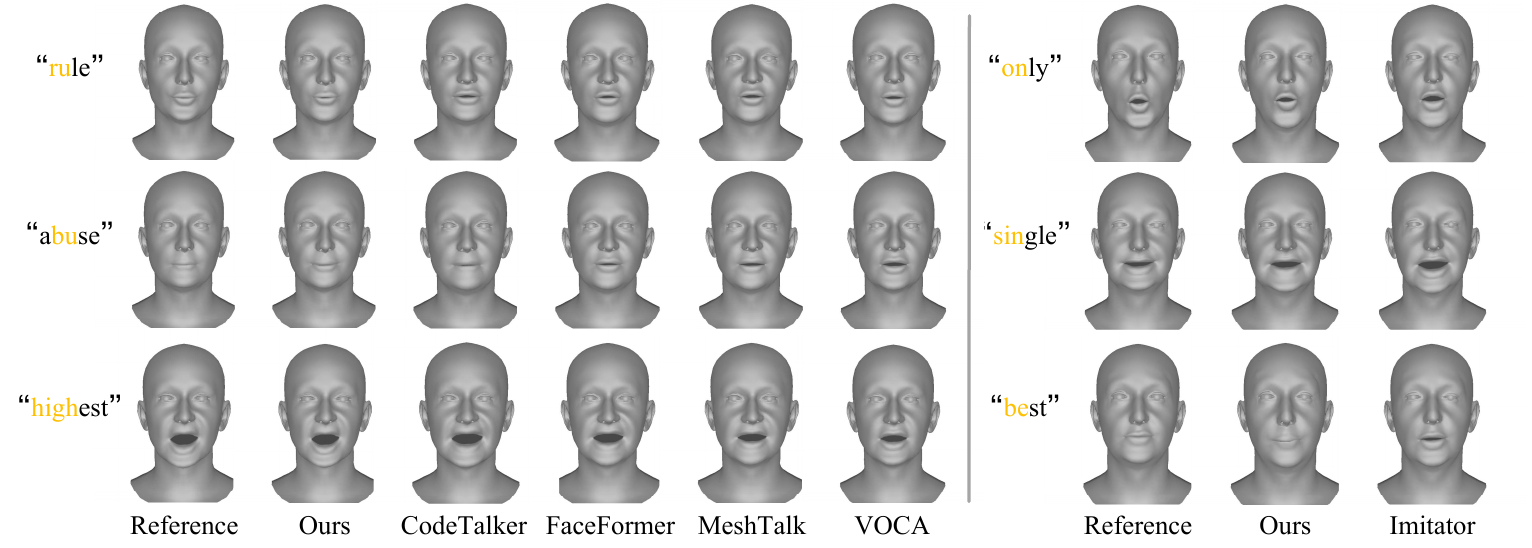}
    \caption{Visual comparisons with state-of-the-art methods on 3D-HDTF Test-A (left) and Test-B (right).}
    \label{render_visual}
\end{figure*}

We visually compare our Mimic with others in Figure \ref{render_visual}. 
Compared to other methods, we achieve results more consistent with the reference ground truth. For example, we achieve accurate lip closures when pronouncing bilabial consonants (e.g., “abuse” and “best”). More importantly, our method can accurately capture the speaking style of the target subject. On the one hand, our method can produce lip shapes with a bit more three-dimensionality in pouting (e.g., “rule” and “only”), chin tucking (e.g., “single”), etc. On the other hand, our method can well learn the amplitude of mouth opening (e.g., “highest”). More remarkable animation comparisons are presented in the supplemental video.

\subsubsection{Latent Space Visualization}
To explore our latent spaces, we use t-SNE \cite{tsne} to visualize the style codes and content codes of the subjects in 3D-HDTF-Test-B. For each subject, we select over 40 clips to extract corresponding latent codes and project them to a 2D space. As shown in Figure \ref{latentspace}, each subject is marked with a distinct color. In Figure \ref{latentspace} (a), the style codes of the same subject cluster in the style space, which implies our Mimic learns the identity-specific speaking style. In Figure \ref{latentspace} (b), the content codes of the same subject hardly cluster into a group, suggesting that our content encoder can encode identity-invariant content representations.

\subsubsection{Style Manipulation}
To verify our semantically meaningful style space, we analyze the ability of style interpolation of our style encoder. We select two reference video clips of different subjects in 3D-HDTF-Test-B and extract the style codes using our style encoder. We get two style codes, $s_1$ and $s_2$, which represent two different speaking styles corresponding to large and slight lip movements, respectively. We interpolate a new style code $s^{\prime}=\omega s_1+(1-\omega)s_2$ with a linear weight $\omega$. We generate facial animations of varying $\omega$ and plot the lip distance curves in Figure \ref{styleinter} (top), which demonstrates the smooth transition of lip amplitudes with a linear variation. We visualize a sampled frame of different interpolation results in Figure \ref{styleinter} (bottom), which shows different amplitudes of mouth opening.

\begin{table}[]
    \centering
        \scalebox{0.65}{
\begin{tabular}{@{}lcccccc@{}}
\toprule
\multirow{2}{*}{Competitors} & \multicolumn{2}{c}{3D-HDTF-Test-A}     & \multicolumn{2}{c}{VOCA-Test} & \multicolumn{2}{c}{BIWI-Test-B}    \\ \cmidrule(l){2-7} 
                             & \multicolumn{1}{l}{Lip Sync} & Realism & Lip Sync       & Realism      & Lip Sync          & Realism        \\ \midrule
Ours vs. VOCA                & 95.21                        & 92.32   & 91.02          & 90.58        & 90.05             & 88.42          \\
Ours vs. MeshTalk            & 80.10                        & 81.64   & 82.54          & 79.91        & 85.25             & 83.57          \\
Ours vs. FaceFormer          & 68.54                        & 66.26   & 72.15          & 74.86        & 70.32             & 67.86          \\
Ours vs. CodeTalker          & 63.82                        & 59.56   & 59.83          & 54.65        & 56.58             & 55.44          \\
Ours vs. GT                  & 40.25                        & 44.85   & 43.28          & 44.53        & 44.75             & 42.54          \\ \midrule
\multirow{2}{*}{Competitors} & \multicolumn{6}{c}{3D-HDTF-Test-B}                                                                          \\ \cmidrule(l){2-7} 
                             & \multicolumn{2}{c}{Lip Sync}           & \multicolumn{2}{c}{Realism}   & \multicolumn{2}{l}{Speaking Style} \\ \midrule
Ours vs. Imitator            & \multicolumn{2}{c}{57.20}              & \multicolumn{2}{c}{56.12}     & \multicolumn{2}{c}{58.68}          \\
Ours vs. GT                  & \multicolumn{2}{c}{37.36}              & \multicolumn{2}{c}{39.43}     & \multicolumn{2}{c}{——}             \\ \bottomrule
\end{tabular}
        }
    \caption{User study results on 3D-HDTF-Test-A, VOCA-Test, BIWI-Test-B, and 3D-HDTF-Test-B.}
    \label{userstudy1}
\end{table}


\subsubsection{User Study}
To further fairly compare our method with VOCA, MeshTalk, FaceFormer, and CodeTalker, we evaluate perceptual lip sync and realism on 3D-HDTF-Test-A, VOCA-Test, and BIWI-Test-B. To compare with Imitator, we evaluate perceptual lip sync, realism, and speaking style on 3D-HDTF-Test-B. The A/B testing is used for each comparison. We invite 30 participants to join our study. For 3D-HDTF-Test-A, we randomly select 20 samples in parallel for five methods and the ground truth (GT), resulting in 100 A vs. B pairs. Each pair is judged by at least 3 different participants, and 300 entries are yielded. The same setting is applied to VOCA-Test and BIWI-Test-B to obtain another 600 entries. For 3D-HDTF-Test-B, we divide all ten videos into 10-second clips and randomly select five clips from each video, obtaining 50 samples for each method. We collect 100 A vs. B pairs for evaluating perceptual lip sync and realism as well as 50 GT vs. A vs. B pairs for evaluating speaking style, yielding 450 entries in total. To evaluate the speaking style, we expose the GT video to users and ask them to select one from A vs. B pair that is more consistent in speaking style with the GT. As shown in Table \ref{userstudy1}, participants favor our method over competitors in terms of perceptual lip sync, realism, and consistent speaking style, suggesting that our produced facial animations have superior perceptual quality.

\subsection{Ablation Studies}
\subsubsection{Content Encoder}
For speech-driven facial animation, we only use the style encoder and audio encoder during inference. To study the effect of the content encoder in training, we construct an experiment training our framework without the content encoder $E_c$. As shown in Table \ref{ablationstudy}, we observe all metrics decrease significantly when removing the content encoder. It is mainly because the content space with only audio features contains less motion-related information that hinders the mapping of audio features to facial motions.
\subsubsection{Constraints}
We investigate the impact of removing different constraints and show results in Table \ref{ablationstudy}, suggesting our Mimic performs best with all four constraints. First, removing the auxiliary style classifier ($\mathbf{C}_{s}$) will produce indistinguishable style codes, significantly deteriorating style-related metrics (LDD and SCS). Second, the content-related metrics (FVE, LVE, and LDTW) increase largely without the auxiliary inverse classifier ($\mathbf{C}_{c}$), which may be caused by the identity-related style information remaining in the content space. Third, the removal of the content contrastive loss ($\mathcal{L}_{\text{con}}$) also results in worse content-related metrics due to the domain gap between the audio and motion modalities. Last, our proposed paired latent cycle losses, $\mathcal{L}_{\text{cycle}}^{s}$ and $\mathcal{L}_{\text{cycle}}^{c}$, can effectively boost the performance on style-related metrics and content-related metrics respectively by enhancing the representational capabilities of the latent codes.

\begin{table}[]
    \centering
        \scalebox{0.7}{
        \begin{tabular}{@{}lccccc@{}}
        \toprule
        Method                               & \begin{tabular}[c]{@{}c@{}}FVE $\downarrow$\\ ($\times 10^{-6}\text{mm}$)\end{tabular} & \begin{tabular}[c]{@{}c@{}}LVE $\downarrow$\\ ($\times 10^{-5}\text{mm}$)\end{tabular} & \begin{tabular}[c]{@{}c@{}}LDTW $\downarrow$\\ ($\times 10^{-4}\text{mm}$)\end{tabular} & \begin{tabular}[c]{@{}c@{}}LDD $\downarrow$\\ ($\times 10^{-5}\text{mm})$\end{tabular} & SCS $\uparrow$ \\ \midrule
        \textbf{Full}                        & \textbf{0.551}                                                                & \textbf{3.20}                                                               & \textbf{6.82}                                                                & \textbf{0.89}                                                               & \textbf{0.995} \\
        w/o $E_c$                            & 0.664                                                                         & 4.18                                                                        & 7.98                                                                         & 1.02                                                                        & 0.975          \\
        w/o $C_{s}$                & 0.556                                                                         & 3.51                                                                        & 6.96                                                                         & 0.96                                                                        & 0.985          \\
        w/o $C_{c}$                & 0.620                                                                         & 3.65                                                                        & 7.30                                                                         & 0.90                                                                        & 0.991          \\
        w/o $\mathcal{L}_{\text{con}}$       & 0.635                                                                         & 3.86                                                                        & 7.64                                                                         & 0.91                                                                        & 0.990          \\
        w/o $\mathcal{L}_{\text{cycle}}^{s}$ & 0.558                                                                         & 3.54                                                                        & 7.06                                                                         & 0.94                                                                        & 0.987          \\
        w/o $\mathcal{L}_{\text{cycle}}^{c}$ & 0.633                                                                         & 3.82                                                                        & 7.55                                                                         & 0.92                                                                        & 0.989          \\ \bottomrule
        \end{tabular}
        }
    \caption{Ablation study results on 3D-HDTF-Test-A.}
    \label{ablationstudy}
\end{table}

\section{Conclusion and Discussion}
In this study, we present Mimic for optimizing the speaking style in speech-driven 3D facial animation by disentangling it from facial motions. With a driving speech clip and a short style reference sequence, Mimic produces facial animations that accurately synchronize with speech and match the speaking style of the target subject.  Extensive qualitative and quantitative experiments demonstrate the superiority of our method. We believe that Mimic holds significant potential for practical applications in speech-driven 3D facial animation. 
However, our solution still relies on high-quality 3D face data, which requires high-precision face capture or accurate face reconstruction. One future direction is to reduce the dependencies of high-fidelity 3D faces.

\section{Acknowledgments}
This work was supported in part by the National Key R\&D Program of China under Grant 2021ZD0111601, in part by the National Natural Science Foundation of China (NSFC) under Grants 62276283, in part by Guangdong Basic and Applied Basic Research Foundation under Grants  2023A1515012985, and in part by Basic and Applied Basic Research Special Projects under Grant SL2022A04J01685.

\bibliography{aaai24}

\begin{thebibliography}{27}
\providecommand{\natexlab}[1]{#1}

\bibitem[{Baevski et~al.(2020)Baevski, Zhou, Mohamed, and Auli}]{wav2vec2.0}
Baevski, A.; Zhou, Y.; Mohamed, A.; and Auli, M. 2020.
\newblock wav2vec 2.0: A Framework for Self-Supervised Learning of Speech Representations.
\newblock \emph{Advances in Neural Information Processing Systems}, 33: 12449--12460.

\bibitem[{Bao et~al.(2023)Bao, Zhang, Qian, Xue, Chen, Zhe, and Kang}]{learningviseme}
Bao, L.; Zhang, H.; Qian, Y.; Xue, T.; Chen, C.; Zhe, X.; and Kang, D. 2023.
\newblock Learning Audio-Driven Viseme Dynamics for 3D Face Animation.
\newblock \emph{arXiv preprint arXiv:2301.06059}.

\bibitem[{Chou, Yeh, and Lee(2019)}]{chou2019one}
Chou, J.-c.; Yeh, C.-c.; and Lee, H.-y. 2019.
\newblock One-shot Voice Conversion by Separating Speaker and Content Representations with Instance Normalization.
\newblock \emph{arXiv preprint arXiv:1904.05742}.

\bibitem[{Cudeiro et~al.(2019)Cudeiro, Bolkart, Laidlaw, Ranjan, and Black}]{voca}
Cudeiro, D.; Bolkart, T.; Laidlaw, C.; Ranjan, A.; and Black, M.~J. 2019.
\newblock Capture, Learning, and Synthesis of 3D Speaking Styles.
\newblock In \emph{CVPR}, 10101--10111.

\bibitem[{Edwards et~al.(2016)Edwards, Landreth, Fiume, and Singh}]{jali}
Edwards, P.; Landreth, C.; Fiume, E.; and Singh, K. 2016.
\newblock Jali: An Animator-Centric Viseme Model for Expressive Lip Synchronization.
\newblock \emph{ACM Transactions on Graphics (TOG)}, 35(4): 1--11.

\bibitem[{Fan et~al.(2022)Fan, Lin, Saito, Wang, and Komura}]{faceformer}
Fan, Y.; Lin, Z.; Saito, J.; Wang, W.; and Komura, T. 2022.
\newblock Faceformer: Speech-Driven 3D Facial Animation with Transformers.
\newblock In \emph{CVPR}, 18770--18780.

\bibitem[{Fanelli et~al.(2010)Fanelli, Gall, Romsdorfer, Weise, and Van~Gool}]{biwi}
Fanelli, G.; Gall, J.; Romsdorfer, H.; Weise, T.; and Van~Gool, L. 2010.
\newblock A 3D Audio-Visual Corpus of Affective Communication.
\newblock \emph{IEEE Transactions on Multimedia}, 12(6): 591--598.

\bibitem[{Filntisis et~al.(2022)Filntisis, Retsinas, Paraperas-Papantoniou, Katsamanis, Roussos, and Maragos}]{spectre}
Filntisis, P.~P.; Retsinas, G.; Paraperas-Papantoniou, F.; Katsamanis, A.; Roussos, A.; and Maragos, P. 2022.
\newblock Visual Speech-Aware Perceptual 3D Facial Expression Reconstruction from Videos.
\newblock \emph{arXiv preprint arXiv:2207.11094}.

\bibitem[{Ganin and Lempitsky(2015)}]{grl}
Ganin, Y.; and Lempitsky, V. 2015.
\newblock Unsupervised Domain Adaptation by Backpropagation.
\newblock In \emph{Proceedings of the 32nd International Conference on Machine Learning}, volume~37, 1180–1189.

\bibitem[{Graves et~al.(2006)Graves, Fern{\'a}ndez, Gomez, and Schmidhuber}]{ctcloss}
Graves, A.; Fern{\'a}ndez, S.; Gomez, F.; and Schmidhuber, J. 2006.
\newblock Connectionist Temporal Classification: Labelling Unsegmented Sequences with Recurrent Neural Networks.
\newblock In \emph{ICML}, 369--376.

\bibitem[{Huang and Belongie(2017)}]{huang2017arbitrary}
Huang, X.; and Belongie, S. 2017.
\newblock Arbitrary Style Transfer in Real-time with Adaptive Instance Normalization.
\newblock In \emph{ICCV}, 1501--1510.

\bibitem[{Ji et~al.(2022)Ji, Zhou, Wang, Wu, Wu, Xu, and Cao}]{EAMM}
Ji, X.; Zhou, H.; Wang, K.; Wu, Q.; Wu, W.; Xu, F.; and Cao, X. 2022.
\newblock EAMM: One-Shot Emotional Talking Face via Audio-Based Emotion-Aware Motion Model.
\newblock In \emph{ACM SIGGRAPH 2022 Conference Proceedings}, 1--10.

\bibitem[{Karras et~al.(2017)Karras, Aila, Laine, Herva, and Lehtinen}]{karras2017audio}
Karras, T.; Aila, T.; Laine, S.; Herva, A.; and Lehtinen, J. 2017.
\newblock Audio-driven Facial Animation by Joint End-to-end Learning of Pose and Emotion.
\newblock \emph{ACM Transactions on Graphics (TOG)}, 36(4): 1--12.

\bibitem[{Li et~al.(2017)Li, Bolkart, Black, Li, and Romero}]{flame}
Li, T.; Bolkart, T.; Black, M.~J.; Li, H.; and Romero, J. 2017.
\newblock Learning a Model of Facial Shape and Expression from 4D Scans.
\newblock \emph{ACM Transactions on Graphics (TOG)}, 36(6): 194--1.

\bibitem[{Min et~al.(2021)Min, Lee, Yang, and Hwang}]{saln}
Min, D.; Lee, D.~B.; Yang, E.; and Hwang, S.~J. 2021.
\newblock Meta-StyleSpeech: Multi-Speaker Adaptive Text-to-Speech Generation.
\newblock In \emph{ICML}, 7748--7759.

\bibitem[{Pang et~al.(2023)Pang, Zhang, Quan, Fan, Cun, Shan, and Yan}]{DPE}
Pang, Y.; Zhang, Y.; Quan, W.; Fan, Y.; Cun, X.; Shan, Y.; and Yan, D.-m. 2023.
\newblock DPE: Disentanglement of Pose and Expression for General Video Portrait Editing.
\newblock In \emph{CVPR}, 427--436.

\bibitem[{Peng et~al.(2023)Peng, Wu, Song, Xu, Zhu, Liu, He, and Fan}]{emotalk}
Peng, Z.; Wu, H.; Song, Z.; Xu, H.; Zhu, X.; Liu, H.; He, J.; and Fan, Z. 2023.
\newblock EmoTalk: Speech-Driven Emotional Disentanglement for 3D Face Animation.
\newblock \emph{arXiv preprint arXiv:2303.11089}.

\bibitem[{Radford et~al.(2021)Radford, Kim, Hallacy, Ramesh, Goh, Agarwal, Sastry, Askell, Mishkin, Clark et~al.}]{clip}
Radford, A.; Kim, J.~W.; Hallacy, C.; Ramesh, A.; Goh, G.; Agarwal, S.; Sastry, G.; Askell, A.; Mishkin, P.; Clark, J.; et~al. 2021.
\newblock Learning Transferable Visual Models From Natural Language Supervision.
\newblock In \emph{ICML}, 8748--8763.

\bibitem[{Richard et~al.(2021{\natexlab{a}})Richard, Lea, Ma, Gall, De~la Torre, and Sheikh}]{richard2021audio}
Richard, A.; Lea, C.; Ma, S.; Gall, J.; De~la Torre, F.; and Sheikh, Y. 2021{\natexlab{a}}.
\newblock Audio- and Gaze-driven Facial Animation of Codec Avatars.
\newblock In \emph{WACV}, 41--50.

\bibitem[{Richard et~al.(2021{\natexlab{b}})Richard, Zollh{\"o}fer, Wen, De~la Torre, and Sheikh}]{meshtalk}
Richard, A.; Zollh{\"o}fer, M.; Wen, Y.; De~la Torre, F.; and Sheikh, Y. 2021{\natexlab{b}}.
\newblock MeshTalk: 3D Face Animation from Speech using Cross-Modality Disentanglement.
\newblock In \emph{ICCV}, 1173--1182.

\bibitem[{Thambiraja et~al.(2022)Thambiraja, Habibie, Aliakbarian, Cosker, Theobalt, and Thies}]{imitator}
Thambiraja, B.; Habibie, I.; Aliakbarian, S.; Cosker, D.; Theobalt, C.; and Thies, J. 2022.
\newblock Imitator: Personalized Speech-driven 3D Facial Animation.
\newblock \emph{arXiv preprint arXiv:2301.00023}.

\bibitem[{Van~der Maaten and Hinton(2008)}]{tsne}
Van~der Maaten, L.; and Hinton, G. 2008.
\newblock Visualizing data using t-SNE.
\newblock \emph{Journal of machine learning research}, 9(11).

\bibitem[{Xing et~al.(2023)Xing, Xia, Zhang, Cun, Wang, and Wong}]{codetalker}
Xing, J.; Xia, M.; Zhang, Y.; Cun, X.; Wang, J.; and Wong, T.-T. 2023.
\newblock CodeTalker: Speech-Driven 3D Facial Animation with Discrete Motion Prior.
\newblock In \emph{CVPR}, 12780--12790.

\bibitem[{Zhang et~al.(2021)Zhang, Li, Ding, and Fan}]{hdtf}
Zhang, Z.; Li, L.; Ding, Y.; and Fan, C. 2021.
\newblock Flow-guided One-shot Talking Face Generation with a High-resolution Audio-visual Dataset.
\newblock In \emph{CVPR}, 3661--3670.

\bibitem[{Zhou et~al.(2019)Zhou, Liu, Liu, Luo, and Wang}]{zhou2019talking}
Zhou, H.; Liu, Y.; Liu, Z.; Luo, P.; and Wang, X. 2019.
\newblock Talking Face Generation by Adversarially Disentangled Audio-Visual Representation.
\newblock In \emph{Proceedings of the AAAI conference on artificial intelligence}, volume~33, 9299--9306.

\bibitem[{Zhou et~al.(2020)Zhou, Han, Shechtman, Echevarria, Kalogerakis, and Li}]{makelttalk}
Zhou, Y.; Han, X.; Shechtman, E.; Echevarria, J.; Kalogerakis, E.; and Li, D. 2020.
\newblock MakeItTalk: Speaker-Aware Talking-Head Animation.
\newblock \emph{ACM Transactions On Graphics (TOG)}, 39(6): 1--15.

\bibitem[{Zhou et~al.(2018)Zhou, Xu, Landreth, Kalogerakis, Maji, and Singh}]{visemenet}
Zhou, Y.; Xu, Z.; Landreth, C.; Kalogerakis, E.; Maji, S.; and Singh, K. 2018.
\newblock VisemeNet: Audio-Driven Animator-Centric Speech Animation.
\newblock \emph{ACM Transactions on Graphics (TOG)}, 37(4): 1--10.

\end{thebibliography}
\end{document}